\documentclass[letterpaper, 10 pt, journal, twoside]{IEEEtran}

\usepackage{graphics} %
\usepackage{epsfig} %
\usepackage{amsmath} %
\usepackage[english]{babel}
\usepackage[utf8]{inputenc}
\usepackage{algorithm}
\usepackage{siunitx}
\usepackage{color}
\usepackage{colortbl}
\usepackage{comment}
\usepackage{caption}

\usepackage{booktabs}       %
\usepackage[noend]{algpseudocode}
\usepackage{multirow}

\usepackage[pagebackref=false,breaklinks=true,colorlinks,bookmarks=false]{hyperref}
\usepackage{pifont}
\newcommand{\cmark}{\ding{51}}%

\def\secref#1{Sec.~\ref{#1}}
\def\figref#1{Fig.~\ref{#1}}
\def\tabref#1{Tab.~\ref{#1}}
\def\eqref#1{Eq.~(\ref{#1})}
\def\algref#1{Alg.~\ref{#1}}
\def\etalcite#1{\etal~\cite{#1}}

\definecolor{lightgray}{rgb}{0.83, 0.83, 0.83}

\makeatletter
\usepackage{xspace}
\DeclareRobustCommand\onedot{\futurelet\@let@token\@onedot}
\def\@onedot{\ifx\@let@token.\else.\null\fi\xspace}
\def\eg{e.g\onedot}

\def\etc{etc\onedot} 
 
\def\etal{{et al}\onedot}
\makeatother

\usepackage{algorithm}
\usepackage[noend]{algpseudocode}
\algnewcommand{\algorithmicand}{\textbf{ and }}
\algnewcommand{\algorithmicor}{\textbf{ or }}
\algnewcommand{\OR}{\algorithmicor}
\algnewcommand{\AND}{\algorithmicand}

\begin{document}

\title{CU-Net: LiDAR Depth-only Completion with Coupled U-Net}

\author{Yufei Wang, Yuchao Dai$^*$, Qi Liu, Peng Yang, Jiadai Sun, and Bo Li$^*$ 
\thanks{
This work was partly supported by the National Key Research and Development Program of China (2018AAA0102803) and the National Natural Science Foundation of China (61871325, 62001394, 61901387). \\
$^*$ Corresponding authors: Yuchao Dai and Bo Li.
}
\thanks{All authors are with the School of Electronics and Information, Northwestern Polytechnical University, Xi'an, 710129, China. 
({\tt\footnotesize e-mail: \{wangyufei1951, daiyuchao, changersunjd\}@gmail.com; \{liuqi, yangpeng, libo\}@nwpu.edu.cn.})
}
}

\maketitle

\begin{abstract}
LiDAR depth-only completion is a challenging task to estimate a dense depth map only from sparse measurement points obtained by LiDAR. Even though the depth-only completion methods have been widely developed, there is still a significant performance gap with the RGB-guided methods that utilize extra color images. We find that existing depth-only methods can obtain satisfactory results in the areas where the measurement points are almost accurate and evenly distributed (denoted as normal areas), while the performance is limited in the areas where the foreground and background points are overlapped due to occlusion (denoted as overlap areas) and the areas where there are no available measurement points around (denoted as blank areas), since the methods have no reliable input information in these areas.
Building upon these observations, we propose an effective Coupled U-Net (CU-Net) architecture for depth-only completion.
Instead of directly using a large network for regression, we employ the local U-Net to estimate accurate values in the normal areas and provide the global U-Net with reliable initial values in the overlap and blank areas. The depth maps predicted by the two coupled U-Nets complement each other and can be fused by learned confidence maps to obtain the final completion results. In addition, we propose a confidence-based outlier removal module, which identifies the regions with outliers and removes outliers using simple judgment conditions. The proposed method boosts the final dense depth maps with fewer parameters and achieves state-of-the-art results on the KITTI benchmark. Moreover, it owns a powerful generalization ability under various depth densities, varying lighting, and weather conditions. The code is released at \url{https://github.com/YufeiWang777/CU-Net}.

\end{abstract}


\section{Introduction}

\IEEEPARstart{A}{cquiring} scene geometry using active depth sensors, \eg LiDAR, has multiple applications in robotics~\cite{VodischULGD22} and SLAM~\cite{FrosiM22}. However, due to the limited number of scan beams and angular resolution, existing LiDARs can only provide sparse measurements. Therefore, depth completion methods have been developed to predict dense depth maps from sparse depth measurements. According to whether the color images are utilized for guidance, existing depth completion methods are divided into the RGB-guided methods and the depth-only methods. 
Despite the RGB-guided methods outperforming the depth-only methods by a wide margin in the evaluation metrics, their performance heavily depends on the quality and style of color images, the unpredictable imaging conditions may cause adverse effects on such methods, especially when encountering rain, snow, and night~\cite{ZhengWHS20}.
In this paper, we focus on the depth-only methods. 

\begin{figure}[!t]
	\centering
    \includegraphics[width=0.49\textwidth]{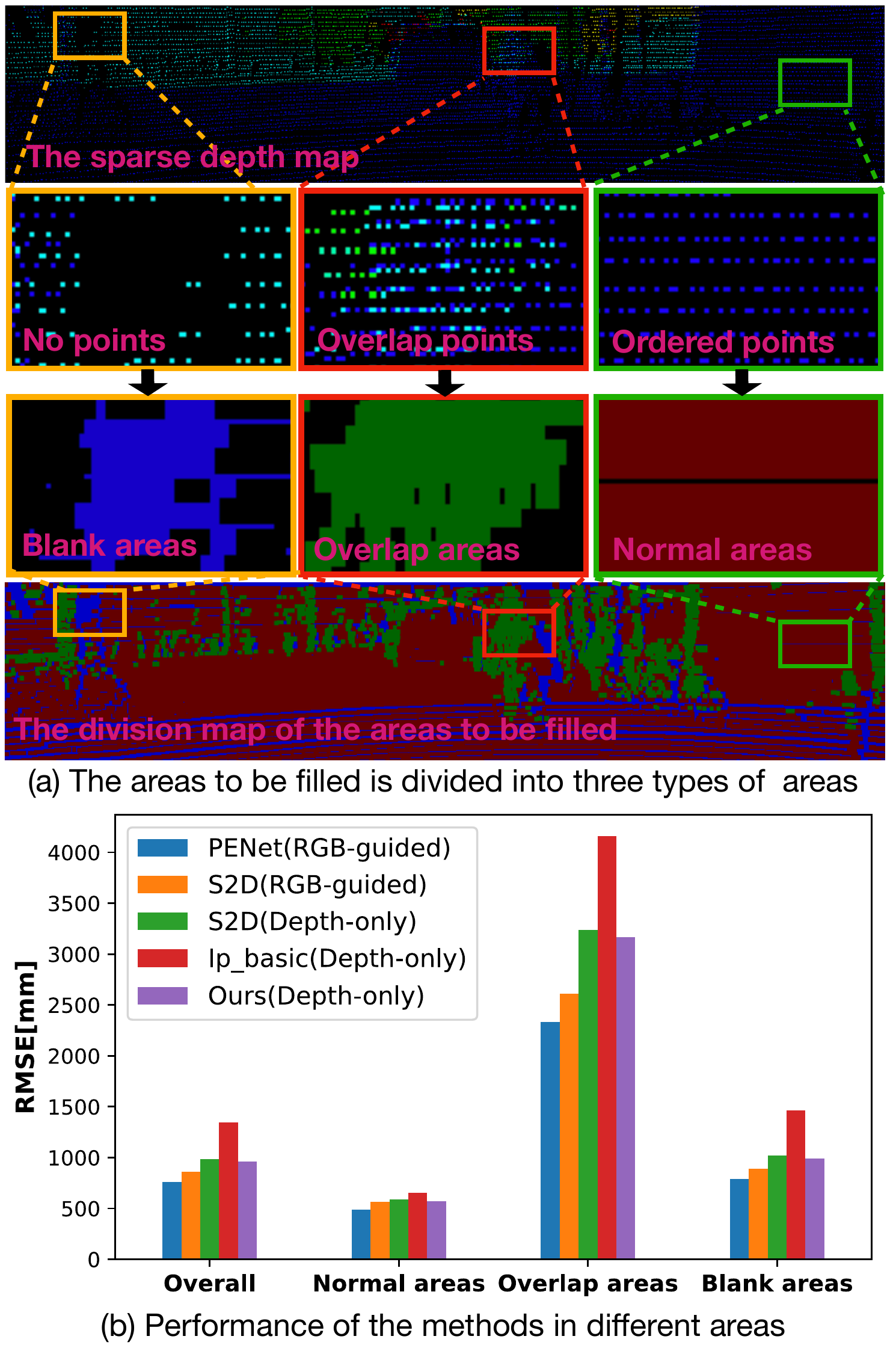}
	\caption{The areas to be filled are divided into normal areas, overlap areas, and blank areas according to different distributions of sparse measurement points.
	Existing depth-only methods can obtain promising results in the normal areas, while in the overlap and blank areas, due to the lack of reliable input information, there is a significant performance gap with the RGB-guided methods. 
	}
    \vspace{-2\baselineskip}
	\label{fig:compared}
\end{figure}

Existing depth-only methods usually employ off-the-shelf network structures to directly regress the sparse depth maps to the dense depth maps, such as sparsity invariant CNN~\cite{UhrigSSFBG17}, encoder-decoder~\cite{MaK18,MaCK19}, and hourglass~\cite{LiYLCZZ20, GansbekeNBG19}. Although the performance of the depth-only methods has been dramatically improved, there is still a large gap with the RGB-guided methods. We observe that there are three different distributions of measurement points in the areas to be filled: 1) the measurement points are almost accurate and the distribution is even, 2) the measurement points are overlapped by foreground and background points due to occlusion, and some of the points whose depth values abruptly change should be removed as outlier points~\cite{Uncertainty, Surface}, 3) there are no available points around. As shown in~\figref{fig:compared}, according to the distributions, we divide the areas that need to be filled into normal areas, overlap areas, and blank areas. Through experiments, we observe that existing depth-only methods such as S2D~(depth-only)~\cite{MaCK19} and Ip\_basic~\cite{IPBasic} can obtain promising results in normal areas, the performance gap between them and state-of-the-art RGB-guided methods such as PENet~\cite{HuWLNFG21} and S2D~(RGB-guided)~\cite{MaCK19} primarily derives from the overlap and blank areas. We consider that this is because the depth-only methods have no reliable input information in these areas, while the RGB-guided methods can take advantage of the rich semantic information of the color images to get better results.

Therefore, predicting initial input information for the blank areas and overlap areas, and removing the outlier points from the overlap areas are the key to the depth-only methods.
To address the issue, we propose the Coupled U-Net (CU-Net) method, which is enhanced by an outlier removal module. First, we adopt the first U-Net to predict an initial depth map and a corresponding confidence map from the sparse LiDAR measurements. The initial dense depth map has accurate depth values and high confidence in the normal areas, and it can also provide the second U-Net with reliable initial depth values in the overlap and blank areas. Second, we propose a confidence-based outlier removal method. Unlike removing the outliers by judging whether each measurement point meets complex conditions~\cite{Surface}, the proposed method uses the confidence map predicted by the first U-Net to identify the regions with outliers and removes the outliers by a simple judgment condition. Then, the corrected sparse depth map and the initial dense map are fed to the second U-Net to obtain a dense depth map with improved results in the overlap and blank areas and the corresponding confidence map. Since the depth map predicted by the first U-Net shows satisfactory results in the ordered depth points and their local areas, while the depth map predicted by the second U-Net has good results in other global areas, we refer to these two U-Nets as the local U-Net and the global U-Net, respectively. The dense depth maps predicted by the local U-Net and the global U-Net are fused by the confidence maps to obtain the final completion result.

We conduct comprehensive experiments to verify the effectiveness and generalization of our method on the KITTI dataset~\cite{kitti1,kitti2} and DDAD dataset~\cite{DDAD}.
Our contributions can be summarized as follows:
\begin{itemize}
	\item
	We quantitatively analyze the cause of the performance gap between depth-only methods and RGB-guided methods, and show that the primary reason for the limited performance of depth-only methods is their lack of reliable input information in overlap regions and blank areas.
	\item
	To address the issue, we propose a two-stage network with learned intermediate confidence maps, where the first network provides initial depth values of the overlap and blank areas for the second network. Furthermore, we propose a confidence-based outlier removal method to enhance the proposed method, which employs a learned confidence map to identify the areas with outliers and remove them.

	\item	
	Experimental results on the popular KITTI benchmark and the DDAD dataset show that our method achieves state-of-the-art performance among all published papers that employ only depth data during training and inference. Meanwhile, it shows powerful generalization capabilities under different depth densities, changing lighting, and weather conditions.

\end{itemize}

\begin{table*}[!t]
	\centering
	\footnotesize
	\setlength{\tabcolsep}{0.8mm}
	\caption{Performance comparison of typical methods on the KITTI validation dataset. The difficulty of filling varies in different areas, and the main performance gap between the depth-only methods and the RGB-guided methods comes from the overlap areas and blank areas.
    Our method improves all metrics and has fewer parameters compared to S2D~(depth-only). The improvements in overlap and blank areas are more significant.}

	\begin{tabular}{@{}l|c|c|c|c|c|c|c|c|c|c@{}}
		\cline{1-1} \cline{3-11}
		\hline
		Methods       &Params            & Areas         & iRMSE[1/km] $\downarrow$  & iMAE[1/km]$\downarrow$  & RMSE[mm] $\downarrow$   & MAE[mm] $\downarrow$    & Rel $\downarrow$      & $\delta^{1} \uparrow$  &$\delta^{2} \uparrow$ & $\delta^{3} \uparrow$  \\ \hline \hline
		
		\multirow{4}{*}{\begin{tabular}[c]{@{}l@{}}S2D~(RGB-guided)~\cite{MaCK19} \end{tabular}}& \multirow{4}{*}{42.82M}& All areas     & 3.1   & 1.7  & 858.2  & 312.8  & 0.0188 & 99.73\%                 & 99.92\%                 & 99.97\%                 \\
		&& Normal areas   & 2.4   & 1.4  & 564.3  & 251.7  & 0.0159 & 99.89\%                 & 99.97\%                 & 99.99\%                 \\
		&& Overlap areas & 6.3   & 2.8  & 2611.0 & 1299.4 & 0.0478  & 97.26\%              & 99.18\%                 & 99.66\%                 \\
		&& Blank areas   & 3.2   & 1.8  & 891.0  & 318.8  & 0.0196 & 99.72\%              & 99.92\%                 & 99.97\%                 \\ \hline
		\multirow{4}{*}{\begin{tabular}[c]{@{}l@{}}S2D~(depth-only)~\cite{MaCK19} \end{tabular}}& \multirow{4}{*}{42.82M}  & All areas     & 3.7   & 1.7  & 985.1  & 286.5  & 0.0177  & 99.57\%                 & 99.86\%                 & 99.94\%                 \\
		&& Normal areas   & 2.6   & 1.1  & 585.6  & 180.9  & 0.0116  & 99.84\%                 & 99.95\%                & 99.98\%                 \\
		&& Overlap areas & 7.8   & 3.3  & 3237.0 & 1518.7 & 0.0608  & 95.03\%                 & 98.39\%                 & 99.33\%                 \\
		&& Blank areas   & 4.0   & 2.0  & 1023.0 & 312.6  & 0.0202  & 99.55\%                 & 99.86\%                 & 99.94\%                 \\ \hline
		
		\multirow{4}{*}{Ours}& \multirow{4}{*}{34.66M}  & All areas     & 2.8   & 1.0  & 958.8  & 245.3  & 0.0131  & 99.59\%                 & 99.87\%                 & 99.94\%                 \\
		&& Normal areas   & 2.0   & 0.8  & 571.5  & 166.7  & 0.0097  & 99.85\%                 & 99.95\%                & 99.98\%                 \\
		&& Overlap areas & 7.2   & 2.8  & 3168.0 & 1437.4 & 0.0549  & 95.42\%                 & 98.48\%                 & 99.36\%                 \\
		&& Blank areas   & 3.0   & 1.1  & 993.1 & 255.9 & 0.0139  & 99.57\%                 & 99.86\%                 & 99.94\%                 \\ \hline
	\end{tabular}
	\vspace{-\baselineskip}
	\label{tab:areas}
\end{table*}

\section{Related Work}

This section introduces representative works of the RGB-guided methods and the depth-only methods. 

\noindent\textbf{RGB-guided methods.}
The input of the RGB-guided methods includes the sparse depth maps and their corresponding color images. How to fuse the information of these two different modalities is an open problem, a straightforward approach called ``early-fusion'' is to concatenate the depth maps and the color images to form a 4D tensor. Ma~\etalcite{MaCK19} propose a ``later-fusion'' method that extracts the feature of the color images and the sparse depth maps separately, and feeds the fusion features into an Encoder-decoder network. Gansbeke~\etalcite{GansbekeNBG19} propose a method based on color images guidance and uncertainty, which employs two branches and achieves better results. Qiu~\etalcite{QiuCZZLZP19} consider that the color images and depth maps are not strongly correlated, and propose a method that consists of the surface normal guided branch and the RGB guided branch. The proposed method first predicts the surface normal from the color images, and the results of two branches are fused through confidence images to obtain the final dense depth maps. Hu~\etalcite{HuWLNFG21} add the 3D information of the sparse depth maps to the convolution operation. Considering that the existing fusion methods between the color images and depth maps are too simple, Tang~\etalcite{TangTFLT21} propose a guided convolutional network for feature fusion. To address the dense depth maps predicted by end-to-end networks that are blurred at the boundaries of objects, a series of affinity-based spatial propagation methods~\cite{Dyspn,ChengWY18,ChengWGY20,ParkJHLK20,lin2022dynamic} have been proposed.

\noindent\textbf{Depth-only methods.}
The depth-only methods adopt only the given sparse depth maps to predict the corresponding dense depth maps. Many classical methods can only efficiently fill relatively dense depth maps, such as bilateral filtering. To fill highly sparse depth maps, Ku~\etalcite{IPBasic} propose a method that employs traditional image processing techniques, such as morphological transformations and image smoothing. The method has a fast processing speed and its performance exceeds that of contemporary learning methods. Zhao~\etalcite{Surface} also propose a non-learning method based on surface geometry, which is enhanced by an outlier removal algorithm. In the deep learning era, considering the sparsity of the input depth maps, Uhrig~\etalcite{UhrigSSFBG17} propose a sparsity invariant CNN. Cheng~\etalcite{ChengZDJL19} further point out that standard CNN can also handle sparse depth maps as long as the network is deep enough. Then, Ma~\etalcite{MaK18,MaCK19} propose an Encoder-decoder model that uses an encoder network to extract features and a decoder network to recover the resolution. Lv~\etalcite{IRL2} propose a method based on multi-task learning that employs image reconstruction as an auxiliary task. The method is not a pure depth-only completion method, which needs color images in training.

\section{Performance analysis of the depth-only methods for depth completion}
\label{sec:role}

Compared with the RGB-guided methods, the depth-only methods have obvious limitations in evaluation metrics. In this section, we investigate the causes of the limitations by comparing the results of the two methods in different areas. We divide the areas to be filled into three categories according to the distributions of measurement points. To reduce the influence of other factors, we choose a typical depth-only method and RGB-guided method for comparison. These two methods both adopt a single U-Net, and the only difference is that the RGB-guided method also feeds the corresponding color images into the network.

\subsection{Division of the Filled Area}

In this subsection, we use $A$ to denote the filled area and the subtraction ``$-$'' to denote the difference between two areas. Depth completion needs to estimate a dense depth map $A_{all}$ with the same resolution as the sparse depth map. 
We first dilate all the measurement points in the sparse depth map to obtain the areas $A_{sparse}$. Subsequently, we perform the same operation to dilate the outlier points collected by the algorithm~\cite{Surface} to obtain the areas $A_{outlier}$. As follows, we divide the areas to be filled into three categories according to the distributions of measurement points.

\begin{itemize}
	
	\item Normal areas ${A}_{normal}$: the measurement points in these areas are accurate and evenly distributed.
	\begin{equation}
		{A}_{normal} = {A}_{sparse} - {A}_{outlier}.
	\end{equation}
	
	\item Overlap areas ${A}_{overlap}$: the measurement points in these areas are a mixture of accurate points and outlier points.
	\begin{equation}
		{A}_{overlap} = {A}_{outlier}.
	\end{equation}
	
	\item Blank areas ${A}_{blank}$: there are no available measurement points in these areas.
	\begin{equation}
		{A}_{blank} = {A}_{all} - {A}_{normal}  - {A}_{overlap}.
	\end{equation}
\end{itemize}
\vspace{-0.5\baselineskip}

\subsection{RGB-guided vs Depth-only}
\begin{figure*}[!t]
	\centering
	\includegraphics[width=0.85\textwidth]{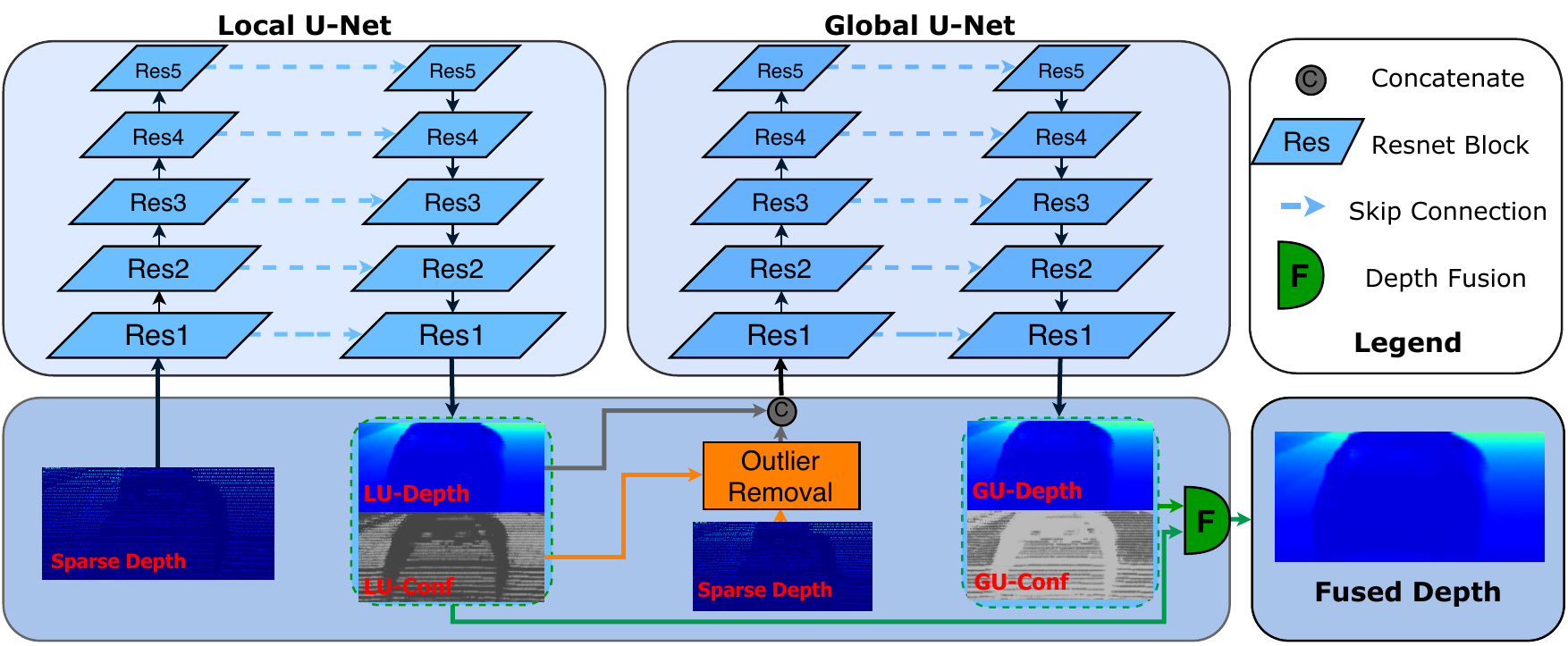}
	\caption{Overview of CU-Net. CU-Net consists of a local U-Net and a global U-Net that have the same structure but do not share parameters. The dense depth map predicted by the local U-Net and the sparse depth map after removing outliers are fed into the global U-Net. CU-Net fuses the dense depth maps predicted by these two U-Nets through learned confidence images to obtain the final results.}
	\vspace{-\baselineskip}
	\label{fig:method}
\end{figure*}
In this subsection, we compare the performance of the typical depth-only method and RGB-guided method in various areas, which are proposed in~\cite{MaCK19} and denoted as S2D~(depth-only) and S2D~(RGB-guided), respectively. For quantitative evaluation, we employ root mean squared error of the inverse depth (iRMSE[1/km]), mean absolute error of the inverse depth (iMAE[1/km]), root mean squared error (RMSE[mm]), mean absolute error (MAE[mm]), relative absolute error (Rel), and $\delta$ inlier ratios (maximal mean relative error of $\delta^{i}=1.25^{i}$ for $i \in {1,2,3}$). Please refer to~\cite{ParkJHLK20} for their specific definitions.

As shown in~\tabref{tab:areas}, there is a significant performance gap between S2D~(RGB-guided) and S2D~(depth-only). We find that the difficulty of filling varies in different areas, and the primary performance gap between the depth-only methods and the RGB-guided methods is reflected in the overlap areas and blank areas. Specifically, S2D~(depth-only) and S2D~(RGB-guided) both achieve satisfactory and comparable performance in the normal areas, where the gap in RMSE is 21.3. However, in the overlap areas and blank areas, S2D~(RGB-guided) obtains better results, where the gaps in RMSE are 626.0 and 132.0, respectively. 
We consider that in the normal areas, whether color images are used or not, the areas to be filled can be well estimated because the measurement points around are accurate and evenly distributed. However, since the blank areas have no valid measurement points and the overlap areas have only overlapped measurement points, it is challenging for the depth-only methods to achieve satisfactory results in these areas due to the lack of reliable input information. On the contrary, RGB-guided methods can exploit the semantic information of color images to improve their performance. In summary, the reason for the limited performance of the depth-only methods is that they have no reliable source of information in the overlap areas and blank areas.

\section{Methods}
\label{sec:method}

To alleviate the limitations of the depth-only methods, we propose the Coupled U-Net model that consists of a local U-Net and a global U-Net. In addition, we propose a confidence-based outlier removal algorithm to remove incorrect points in the sparse depth map before feeding it into the global U-Net. The dense depth maps predicted by these two U-Nets are fused by the learned confidence images to obtain the final dense depth map. In this section, we introduce each module of the proposed method in detail. The proposed network architecture is shown in~\figref{fig:method}. 

\subsection{Coupled U-Net Model}

Existing depth-only methods treat depth completion as an image-to-image task. Given a sparse depth map $D_{\text {sparse}}$, they usually use an end-to-end network such as U-Net to learn the mapping from the sparse depth map $D_{\text {sparse}}$ to the dense depth map $D_{\text {dense}}$. As shown in \eqref{equ:sunet}, U-Net consists of an encoder network and a decoder network. The encoder network $f_{\text {encoder}}$ extracts the feature representation $X$ from the sparse depth map $D_{\text {sparse}}$. Then, the decoder network $f_{\text {decoder}}$ uses operations such as deconvolution or upsampling to predict a dense depth map $\hat{D}_{\text {dense}}$ with the same resolution as $D_{\text {sparse}}$.
\begin{equation}
\hspace{-1pt}
	\! \hat{D}_{\text{dense}} \! = \! f_{\text{U-Net}}\left(D_{\text{sparse}}\right) \! \Longrightarrow \! \left\{\begin{array}{l}
		X = f_{\text{encoder}}\left( D_{\text{sparse}}\right)  \\
		\hat{D}_{\text{dense}} = f_{\text{decoder}}\left(X\right)
	\end{array}\right.. \label{equ:sunet}
\end{equation}

As discussed in \secref{sec:role}, a single U-Net-based method can obtain satisfactory results in the normal region. However, there are undesirably large errors in the overlap and blank areas due to the lack of reliable input information. To solve this issue, we first use the local U-Net to predict an initial depth map, and then employ another U-Net, the global U-Net, to further enhance the results in the overlap areas and blank areas. Different from existing depth-only methods, the local U-Net also estimates a confidence map $\hat{C}_{LU}$ that reflects the reliability of the initial depth map. The local U-Net is described in \eqref{equ:localunet}.
\begin{equation}
	(\hat{D}_{LU}, \hat{C}_{LU}) = f_{\text {LocalUNet}}\left(D_{\text {sparse}}\right).
	\label{equ:localunet}
\end{equation}

Although the dense depth map predicted by the local U-Net has unsatisfactory results in the overlap areas and blank areas, it is adequate to provide better initial values than the sparse depth maps. Consequently, we employ it to guide the global U-Net to generate better results in these areas. We feed the dense depth map $\hat{D}_{LU}$ and the sparse depth map $ D_{\text {sparse}}$ into the global U-Net. The global U-Net also predicts a dense depth map $\hat{D}_{GU}$ and the corresponding confidence map $\hat{C}_{GU}$. The global U-Net is described in \eqref{equ:globalunet}.
\begin{equation}
	(\hat{D}_{GU}, \hat{C}_{GU}) = f_{\text{GlobalUNet}}\left(\hat{D}_{LU}, D_{\text{sparse}}\right).
	\label{equ:globalunet}
\end{equation}

\subsection{Outlier Removal}
\label{subsec:outlier_removal}

In this section, we propose a confidence-based outlier removal method to remove incorrect points of the sparse depth maps. The algorithm~\cite{Surface} removes outlier points by judging whether each point in the sparse depth map meets the set of complex conditions. For example, the outlier points may change the relative position with some surrounding correct points, have relatively larger values, or create denser areas. However, due to the irregular distributions of points in the sparse depth map, it will also remove some correct points that meet the conditions, resulting in the loss of information. Unlike the algorithm~\cite{Surface} which works on the whole sparse depth map, we propose a confidence-based outlier removal method that focuses on the overlap areas of the sparse depth maps, where the outlier points are mainly clustered. Since the confidence maps predicted by the local U-Net have high values in the normal areas and low values in the overlap and blank areas, and there are no points in the blank areas, we can employ them to identify the overlap areas by setting a confidence threshold. Since the outlier points usually arise from the further background, the proposed method can achieve effective outlier removal by removing the points whose depth values are larger than the surrounding points, thus retaining more useful information.

The proposed outlier removal algorithm is shown in \algref{alg:algorithm}. $\mathbf{N}_d$ denotes the nonzero points set, which consists of the neighborhood points of a measurement point $d$, and $|\mathbf{N}_d|$ is the cardinal number. A measurement point $d$ is deemed valid, only if two conditions are met. First, the corresponding confidence $\hat{C}_{LU}(u, v)$ exceeds a threshold $\epsilon$. Second, the difference between the measurement point and the average of its nonzero neighborhood is less than the threshold $\varepsilon$. The valid measurement $d$ is added to $D_r$, which will be concatenated with $\hat{D}_{LU}$.

\begin{figure}[!t]
\begin{algorithm}[H]%
	\caption{The proposed outlier removal algorithm}
	\label{alg:algorithm}
	\textbf{Initialization:\! $D_{r}\leftarrow\emptyset$}
	\begin{algorithmic}[0] 
		\For{ $d \in D_{sparse}$ \AND $d \neq 0$} 
		\State $(u, v) \leftarrow \textbf{Index}(d)$
		\If{$\hat{C}_{LU}(u, v) > \epsilon$ \OR $|d - \dfrac{1}{|\mathbf{N}_d|}\sum\limits_{d_i \in \mathbf{N}_d} d_i| < \varepsilon$}
		\State $D_{r} = D_{r} \cup \{d\}$ 
		\EndIf
		\EndFor
		\Return $D_{r}$
	\end{algorithmic}
\end{algorithm}
\vspace{-1.5\baselineskip}
\end{figure}

\subsection{Depth Fusion by Confidence}

The dense depth maps predicted by the local U-Net $\hat{D}_{LU}$ have accurate values in the normal areas and coarse values in the overlap areas and blank areas, while the dense depth maps predicted by the global U-Net $\hat{D}_{GU}$ have the opposite. It indicates that these two dense depth maps have unique and complementary attributes. To take advantage of their respective strengths, we fuse them through the learned confidence maps $\hat{C}_{LU}$ and $\hat{C}_{GU}$.
The fused depth $\hat{D}$ is obtained by \eqref{equ:fusion}, $(u,v)$ denotes the coordinate of the pixel.
\begin{equation}
	\hat{D}(u, v)=\frac{{\hat{C}_{LU}(u, v)} \cdot \hat{D}_{LU}(u, v)+{\hat{C}_{GU}(u, v)} \cdot \hat{D}_{GU}(u, v)}{{\hat{C}_{LU}(u, v)}+{\hat{C}_{GU}(u, v)}}.
	\label{equ:fusion}
	\vspace{-1\baselineskip}
\end{equation}

\subsection{Loss Function}

We use an L2 loss that is defined in \eqref{equ:l2}.
\begin{equation}
	\mathcal{L}(\hat{D})=\| (\hat{D} - D_{gt}) \odot \mathbf{1}_{\{D_{gt} > 0\}}  \|^{2}. \label{equ:l2}
\end{equation}
where $\hat{D}$ is the predicted dense map, $D_{gt}$ is the ground truth for supervision. Since the ground truth depth maps are semi-dense, which are created by aggregating 11 consecutive LiDAR scans into one, we only supervise the valuable parts, $\mathbf{1}$ indicates whether there is a value in the ground truth, $\odot$ denotes the element-wise multiplication. In training, we supervise the dense depth map predicted by the local U-Net $\hat{D}_{LU}$, the dense depth map predicted by the global U-Net $\hat{D}_{GU}$, and the fused dense depth map $\hat{D}$ simultaneously. The two confidence maps $\hat{C}_{LU}$ and $\hat{C}_{GU}$ are learned without direct supervision. The final loss is defined in \eqref{equ:loss}, where $\lambda_{LU}$, $\lambda_{GU}$, and $\lambda_{Fused}$ are three hyper-parameters.
\begin{equation}
	\mathcal{L} = \lambda_{LU} \mathcal{L}(\hat{D}_{LU}) + \lambda_{GU}\mathcal{L}(\hat{D}_{GU}) + \lambda_{Fused} \mathcal{L}(\hat{D}).
	\label{equ:loss}
\end{equation}

\section{Experiments}

We perform extensive experiments to verify our method. We first introduce the training setting and the datasets in \secref{sec:training}. Then, we compare our method with the state-of-the-art depth-only methods on the KITTI dataset in \secref{sec:sota}. In addition, we conduct comprehensive ablation experiments to verify the effectiveness of each component of the proposed method in \secref{sec:ablation}. Last but not least, in \secref{sec:discuss}, we perform experiments in various settings including different densities of the input depth, changing lighting, and weather conditions to prove the generalizability of our method.

\begin{figure*}[!t]
	\centering
	\includegraphics[width=0.98\textwidth]{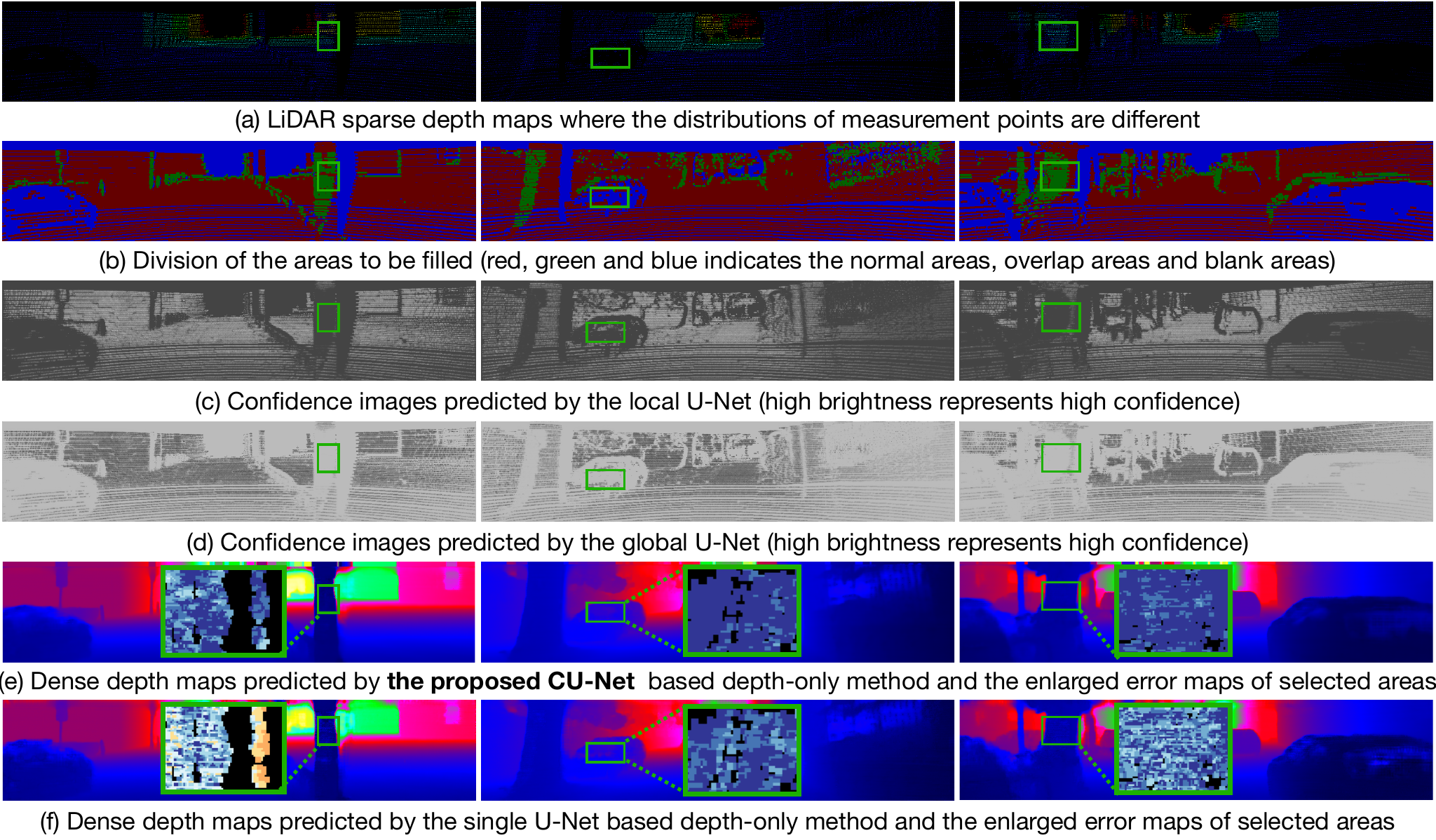}
	\vspace{-2mm}
	\caption{Depth completion results on the KITTI dataset. (a) the sparse input depths; (b) the division of the area to be filled; (c) and (d) are confidence images predicted by the local U-Net and the global U-Net; (e) and (f) are the predicted dense depth maps of our method and S2D (depth-only)\cite{MaCK19}, the error maps of corresponding areas are enlarged.}
	\vspace{-3mm}
	\label{fig:result}
\end{figure*}

\subsection{Experiment Setups}
\label{sec:training}

We use PyTorch to implement our method, which is trained with a single Tesla V100 GPU. During training, we set the batch size to 6 and use a step learning rate decay policy with the Adam optimizer. The initial learning rate is set to $10^{-3}$, and we reduce the learning rate by 50\% every 5 epochs. We set $\beta_{1} = 0.9$, $\beta_{2} = 0.999$, weight decay is $10^{-6}$, and the total number of epochs is 15. In outlier removal, we select a $7\times7$ window as the neighborhood, $\epsilon$ and $\varepsilon$ are set to $0.7$ and $1$, respectively. The joint training hyper-parameters $\lambda_{LU}$, $\lambda_{GU}$, and $ \lambda_{Fused}$ in \eqref{equ:loss} are set to 0.3, 0.3, and 0.5, respectively. We use the same ResNet block as in~\cite{HuWLNFG21}.

\textbf{KITTI Dataset.} The KITTI dataset~\cite{kitti1,kitti2} is a large autonomous driving real-world dataset, which consists of sparse depth maps obtained by raw LiDAR scans and corresponding color images. The dataset contains 86k training data, 1k selected validation data, and 1k test data without ground truth, which needs to be evaluated on the KITTI online benchmark server. Since the top of the sparse depth map does not have valid LiDAR points, we crop the input images to $256\times1216$ for both training and inference as~\cite{TangTFLT21}.

\textbf{DDAD Dataset.} The Dense Depth for Autonomous Driving (DDAD) dataset~\cite{DDAD} is a new autonomous driving benchmark from the Toyota Research Institute for long-range (up to 250m).
The dataset contains monocular color images and accurate ground-truth depth generated from a high-density, long-range Luminar-H2 LiDAR sensor. Unlike the KITTI dataset captured under good lighting and weather conditions, the DDAD dataset contains scenes with different lighting conditions (such as morning, sunset, and overcast) and weather conditions (for example rain).  
We conduct experiments under various conditions in the DDAD dataset to verify the generalizability of our model. In experiments, we crop the input depth maps to $512\times1216$.

\subsection{Comparison with state-of-the-art Methods}
\label{sec:sota}

We evaluate our method on the KITTI depth completion. We train our model end-to-end \textbf{from scratch} on the train set and compare the performance with other top-ranking depth-only completion methods on the KITTI leaderboard. As a depth-only solution, our method employs \textbf{only depth data} during training and inference. So we first compare it with the methods with the same settings, including non-learning methods IP\_Basic~\cite{IPBasic} and Physical\_Surface~\cite{Surface}, as well as learning methods SparseConvs~\cite{UhrigSSFBG17}, Nconv\_CNN~\cite{NconvCNN}, pNCNN~\cite{pNCNN}, \etc. As shown in~\tabref{tab:sota}, our method outperforms all other methods in the main RMSE metric and achieves the best performance in MAE, iRMSE, and iMAE metrics.

Furthermore, we compare our method with other methods that use extra information during training.
For example, Spade-sD~\cite{SpadesD} uses the synthetic data, Glob\_guide~\cite{Globguide} uses semantic segmentation labels, and IR\_L2~\cite{IRL2} needs RGB as auxiliary information. Compared with these methods, our method still achieves comparable results. Specifically, our method achieves better results than Glob\_guide~\cite{Globguide} across all four metrics. Meanwhile, it has smaller RMSE metrics than Spade-sD~\cite{SpadesD} and smaller MAE, iRMSE, and iMAE than IR\_L2~\cite{IRL2}.

\begin{table}[!t]
	\centering
	\setlength{\tabcolsep}{1.4 mm}
	\caption{Quantitative comparison with state-of-the-art depth-only completion algorithms on the KITTI leaderboard. Whether the methods use extra information in the training is indicated in the column ``ExInfo''.}
	\begin{tabular}{@{}l|c|c|c|c|c@{}}
		\hline
		Methods     & iRMSE$\downarrow$       & iMAE$\downarrow$       & RMSE$\downarrow$         & MAE$\downarrow$             & ExInfo      \\ \hline \hline
		Spade-sD~\cite{SpadesD}    & \textbf{2.60 }         & \textbf{0.98 }         & 1035.29         & \textbf{248.32 }         & Synthetic   \\ 
		Glob\_guide~\cite{Globguide} & 2.80          & 1.07          & 922.93          & 249.11          & Semantic    \\ 
		IR\_L2~\cite{IRL2}      & 4.92          & 1.35          &\textbf{ 901.43 }         & 292.36          & RGB         \\ \hline
		SparseConvs~\cite{UhrigSSFBG17} & 4.94          & 1.78          & 1601.33         & 481.27          & No          \\ 
		IP\_Basic~\cite{IPBasic} 		&3.78			&1.29			&1288.46			&302.60			& No			\\
		ADNN~\cite{ADNN}        & 59.39         & 3.19          & 1325.37         & 439.48          & No          \\ 
		Nconv\_CNN~\cite{NconvCNN}  & 4.67          & 1.52          & 1268.22         & 360.28          & No          \\ 
		S2D~(depth-only)~\cite{MaCK19}      & 3.21          & 1.35          & 954.36          & 288.64          & No          \\ 
		HMS-Net~\cite{HMSNet}     & 2.93          & 1.14          & 937.27          & 258.48          & No          \\ 
		pNCNN~\cite{pNCNN}       & 3.37          & 1.05          & 960.05          & 251.77          & No          \\ 
		Physical\_Surface~\cite{Surface}     & 3.76          & 1.21          & 1239.84         & 298.30          & No          \\ 
		DTP~\cite{Pooling}     & 2.94          & 1.07          & 937.27          & 247.81          & No          \\  \hline
		\textbf{Ours}        & \textbf{2.69} & \textbf{1.04} & \textbf{917.76} & \textbf{244.36} & \textbf{No} \\ \hline
	\end{tabular}
    \vspace{-\baselineskip}
	\label{tab:sota}
\end{table}

\subsection{Ablation Study}
\label{sec:ablation}

\begin{table*}[!t]
	\centering
	\caption{Performance of different variants obtained by varying the input of global U-Net on the KITTI validation dataset. The best results are in bold, and our method, the results are highlighted with a gray background, achieves the minimum RMSE.}
	\vspace{-0.5\baselineskip}
	\begin{tabular}{@{}c|cccc|c|c|c|c|c|c|c|c@{}}
		\hline
		Models & SD 			 & SCD			 & LU-Depth 			  & LU-Guidance 	   & iRMSE $\downarrow$        & iMAE $\downarrow$  	  & RMSE $\downarrow$      & MAE $\downarrow$       & Rel $\downarrow$   & $\delta^{1} \uparrow$ 	& $\delta^{2} \uparrow$ 		&$\delta^{3} \uparrow$ \\ \hline \hline
		$M_1$ & \cmark	  &     				&    				&       			& 3.4          		& 1.2        		& 962.9           & 260.7           & 0.0146 	 & 99.59\%               	& 99.87\%                & \textbf{99.95\%}              \\ 
		$M_2$ &     				&     				& \cmark	&       			& 2.9          		& 1.1        		& 959.8           & 254.9           & 0.0138 	  & \textbf{99.60\%}           		& 99.87\%                & 99.94\%               \\ 
		$M_3$ &     				&     				&    				&\cmark    & 2.9          		& 1.1        		& 960.4           & 258.0           & 0.0142 	  & 99.59\%                  & 99.87\%                & 99.94\%               \\ 
		$M_4$ & \cmark	   	&     				&\cmark  &       				& \textbf{2.7}		  		& 1.0				& 963.6           & \textbf{242.5} 			& \textbf{0.0129}	  & 99.60\%                  & 99.87\%                & 99.94\%               \\ 
		 \rowcolor{lightgray} $M_5$ &     				&\cmark	&\cmark	&       			& {2.8}         		& {\textbf{1.0}}        	    & {\textbf{958.8}}			  & {245.3}           & {0.0131}      & {99.59\%}                  & {\textbf{99.87\%}}               & {\textbf{99.94\%}}               \\ 
		$M_6$ &     				&\cmark	&    				&\cmark	& 2.8          		& 1.1         		& 959.8           & 249.2           & 0.0133      & 99.60\%                   & 99.87\%               & 99.95\%               \\ \hline
	\end{tabular}
	\vspace{-0.5\baselineskip}
	\label{tab:ablation}
\end{table*}

\subsubsection{\textbf{Coupled U-Net Model}}

We first compare our method with the typical depth-only methods and RGB-guided methods, both of which adopt a single U-Net. The results in~\tabref{tab:areas} show that our method improves all metrics compared with the typical depth-only methods, and the improvement in the areas of overlap and blanks is more significant. Since the overlap area is mainly concentrated on the boundary of the object, which is difficult for the depth-only method, the result of the overlap area is still far from that of other areas. Meanwhile, it is worth mentioning that our method adopts a more light-weight U-Net, so the network parameters are fewer.

By varying the input of the global U-Net, our method has six different variants, denoted by $M_1$ to $M_6$. The detailed configuration and comparison results of these six variants are shown in~\tabref{tab:ablation}, where SD denotes the sparse depth map, SCD denotes the sparse depth map after removing outlier points, and LU-Depth denotes the dense depth map predicted by the local U-Net. The results show that the setup of our method achieves the minimum RMSE that is the basis for the ranking of the KITTI leaderboard. In addition, we also verify the effect of employing an extra guidance image that is applied to guide the global U-Net (denoted by LU-Guidance), as done in~\cite{GansbekeNBG19} and~\cite{QiuCZZLZP19}. The results show that the additional guidance maps do not provide any performance benefits.

Some typical examples are shown in~\figref{fig:result}, which show that the confidence images predicted by the local U-Net have high confidence values in the normal areas, whereas they have low confidence values in the overlap areas and blank areas. The confidence images predicted by the global U-Net are the opposite. This indicates the dense depth maps predicted by the local U-Net and the dense depth maps predicted by the global U-Net are complementary. We also make a visual comparison between the dense depth maps predicted by our proposed method and the dense depth maps predicted by the typical depth-only method~\cite{MaCK19}. From the enlarged error maps, we can see that our method has a smaller error in the overlap and blank areas, which shows the effectiveness of our method. 

\begin{figure}[!t]
	\centering
	\includegraphics[width=0.48\textwidth]{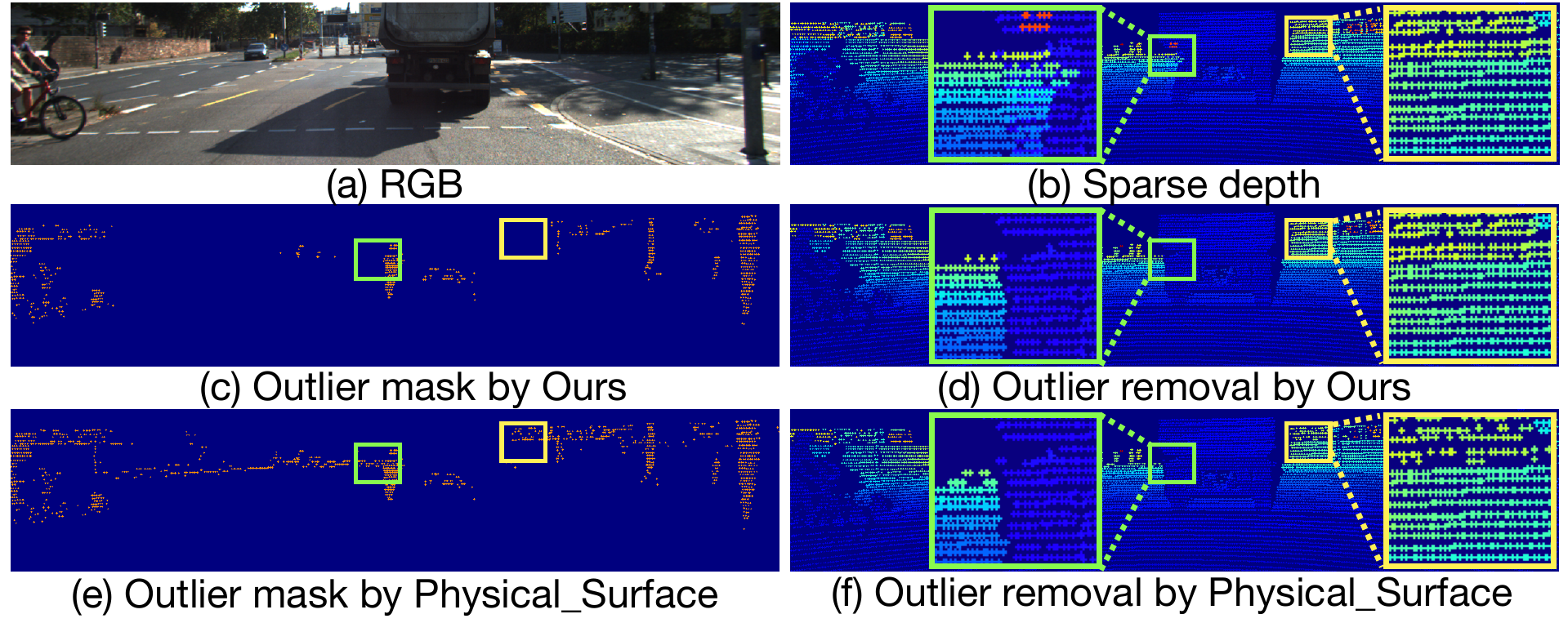}
	\vspace{-0.5\baselineskip}
	\caption{Examples of the outlier removal.
	(a) the RGB images; (b) the sparse input depths; (c) and (d) are the outlier mask predicted by our method and the depth map after removing outlier points; (e) and (f) are the outlier mask predicted by Physica\_Surface~\cite{Surface} and the depth map after removing outlier points.}
	\label{fig:outlier}
    \vspace{-\baselineskip}
\end{figure}

\subsubsection{\textbf{Outlier Removal}}

To reduce the correct points removed by mistake, we propose a confidence-based outlier removal method that first identifies the areas where the outlier points may exist, and then removes outliers from the areas by simple judgment conditions. We select the outlier removal method~\cite{Surface} that works on the entire sparse depth map as the baseline. To verify the effectiveness of the proposed method, we quantitatively evaluate the quality of the sparse depth maps by comparing them with the ground truth. In this process, we only consider the points where the sparse depth map and the ground truth both have values. As shown in~\tabref{tab:outlier}, the original sparse depth maps have a significant error. Our method and the baseline method effectively improve the quality of the sparse depth maps by removing the outlier points. However, compared with the baseline method, our method obtains better results in all metrics and retains 95.34\% of the points in the sparse depth map. In contrast, the baseline method retains 94.00\% of the points in the sparse depth map. \figref{fig:outlier} also shows that our method effectively removes erroneous points in the sparse depth maps while retaining more measurement points.

In addition, we compare the RMSE of the proposed CU-Net using our confidence-based outlier removal method and the baseline method~\cite{Surface} in~\tabref{tab:outlier2}. The RMSE using the proposed method is 958.8 while the RMSE of the baseline method is 964.36. It also shows the effectiveness of our method.

\begin{table}[!th]
	\centering
	\setlength{\tabcolsep}{0.7mm}
	\caption{Performance comparison with different outlier removal algorithms on the KITTI validation dataset.
	}
	\begin{tabular}{@{}l|c|c|c|c|c@{}}
		\hline
		Depths          & iRMSE $\downarrow$  & iMAE $\downarrow$  & RMSE $\downarrow$   & MAE $\downarrow$    & keep ratio $\uparrow$  \\ \hline \hline
		Raw depth map  & 5.0   & 0.8  & 1595.2 & 202.5 & 100.00\%   \\ 
		Baseline method~\cite{Surface}   & 2.1   & 0.4  & 662.9  & 81.7  & 94.00\%    \\ \hline
		\textbf{Ours}            & \textbf{1.7}   & \textbf{0.3}  & \textbf{466.1}  & \textbf{73.2}  & \textbf{95.34\%}   \\
		\hline
	\end{tabular}
	\label{tab:outlier}
\end{table}

\begin{table}[!t]
	\centering
	\setlength{\tabcolsep}{0.7mm}
	\caption{Ablation experiments of the outlier removal method on the KITTI validation dataset.
	}
	\begin{tabular}{@{}l|c|c|c|c@{}}
		\hline
		Algorithms          & iRMSE $\downarrow$  & iMAE $\downarrow$  & RMSE $\downarrow$    & MAE $\downarrow$     \\ \hline \hline
		Baseline method~\cite{Surface} & 2.8  & 1.0  &964.36  & 346.30    \\  \hline
		\textbf{Ours}  & \textbf{2.8}   & \textbf{1.0}  &\textbf{958.8} & \textbf{245.3}    \\
		\hline
	\end{tabular}
	\label{tab:outlier2}
\end{table}

\subsection{Generalization Capability} 
\label{sec:discuss}

To prove the generalization capability of our method, we evaluate its performance under different depth densities, changing lighting, and weather conditions.

\subsubsection{\textbf{Generalization on more sparse depth maps}}

The sparse depth map of the KITTI dataset is obtained by a 64-line Velodyne LiDAR. However, in many practical applications, only 32-line or 16-line LiDAR will be employed due to cost constraints, which only provide more sparse depth maps. Therefore, it is crucial to analyze the performance of the method on sparse depth maps with different sparsity levels.

\begin{figure}[!t]
	\centering
	\includegraphics[width=0.45\textwidth]{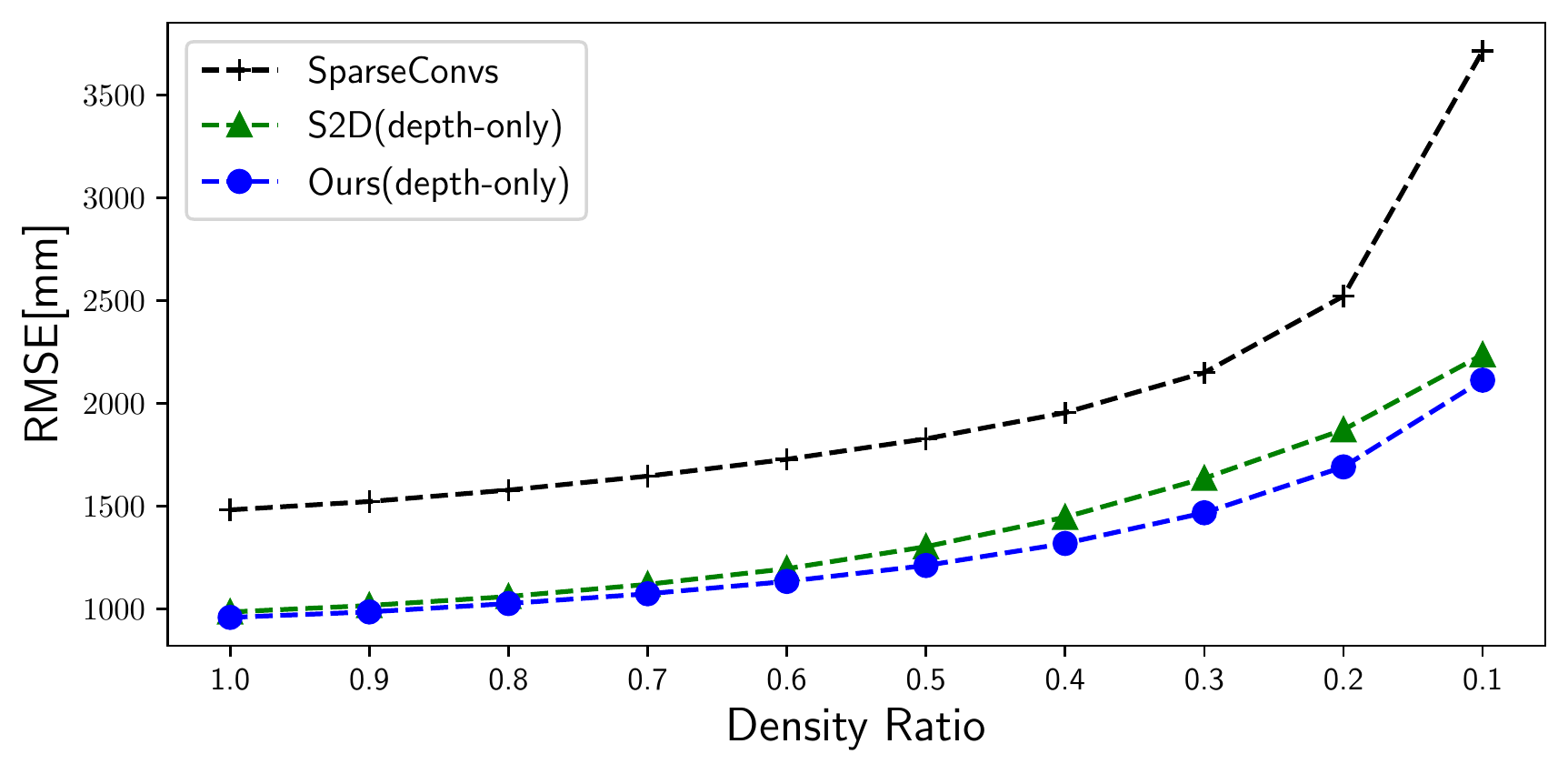}
	\vspace{-0.5\baselineskip}
	\caption{The performance of the methods under different depth densities.}
	\label{fig:density}
	\vspace{-\baselineskip}
\end{figure}

We use the depth map obtained by the 64-line LiDAR as the baseline. Meanwhile, we employ the random sampling strategy to generate more sparse depth maps according to different ratios of density. In~\figref{fig:density}, we compare the performance of our method with S2D~(depth-only)~\cite{MaCK19} and SparseConvs~\cite{UhrigSSFBG17} under different sparsity levels on the KITTI validation dataset. To ensure the fairness of the comparison, our model is trained only on the KITTI train dataset without fine-tuning on depth maps with different sparsity levels, while the results of S2D~(depth-only) are obtained by open-source code. \figref{fig:density} shows that the performance of the model gradually decreases with the depth density decreasing. However, our method outperforms other methods at different density ratios, which shows that our model has better generalization capability on more sparse depth maps.

\subsubsection{\textbf{Generalization under various conditions}}

\begin{figure}[!t]
	\centering
	\includegraphics[width=0.45\textwidth]{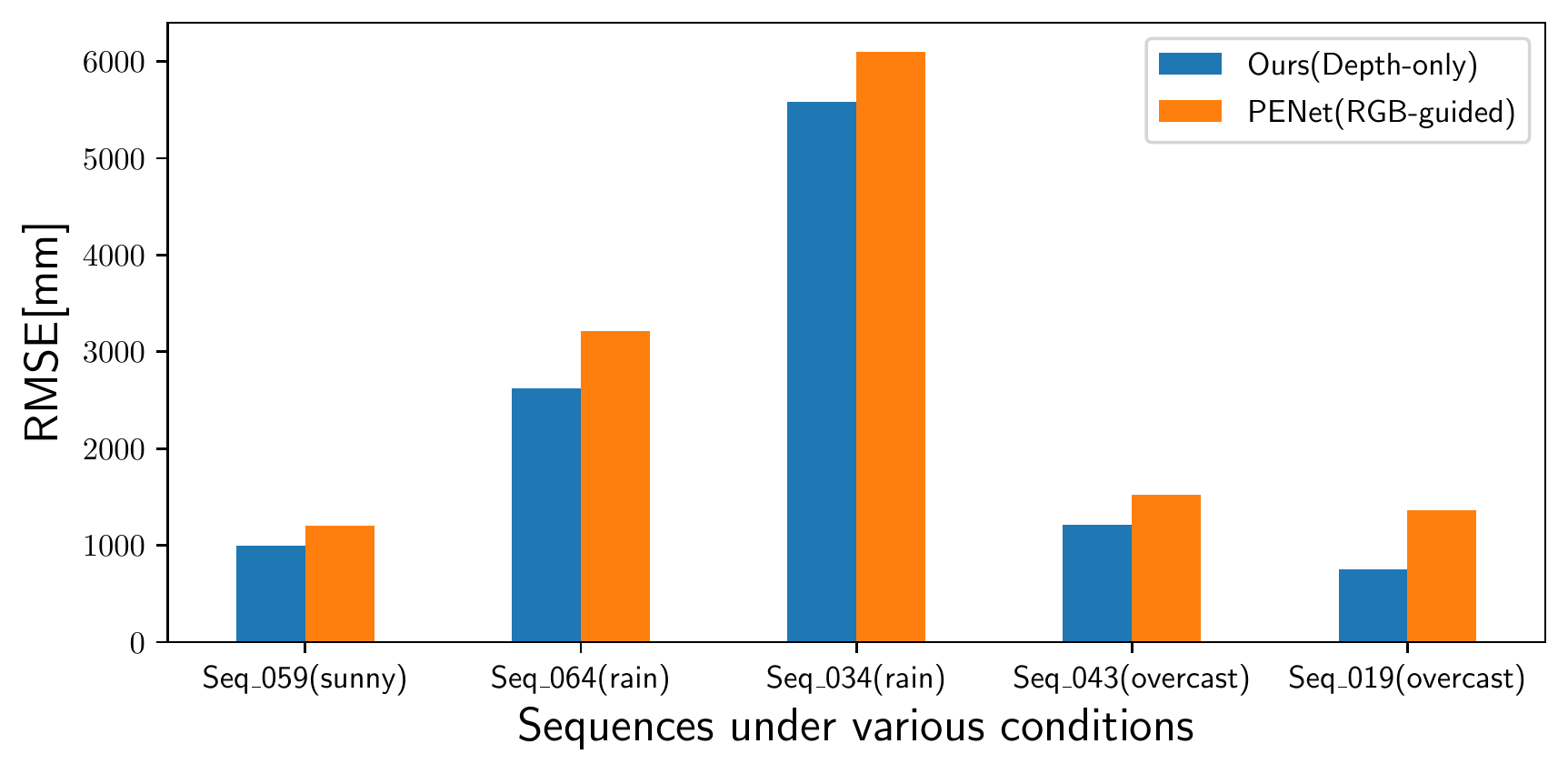}
	\caption{The performance under the sequences with different lighting and weather conditions on the DDAE dataset.} \label{fig:conditions}
	\vspace{-\baselineskip}
\end{figure}

As a depth-only solution, our method has stronger robustness under different conditions, as it is not affected by the quality and style of color images. To verify the generalization ability of our model in various conditions and datasets, we use the model trained on the KITTI dataset to directly test on five sequences with different lighting and weather conditions of the DDAD dataset. The experimental setup is difficult because there are many differences between training and inference data. We compare our method with the state-of-the-art RGB-guided method PENet~\cite{HuWLNFG21}. The PENet model is provided by the open-source project, whose RMSE is 757 in the KITTI validation dataset. As shown in~\figref{fig:conditions}, compared with PENet, our method achieves better results in the five sequences and is relatively stable under changing conditions. It shows that our method has good generalization ability under various conditions.

\vspace{-0.2cm}
\section{Conclusions}

By investigating the reasons for the performance gap between the depth-only completion methods and the RGB-guided completion methods, we observed that the lack of reliable input information in the overlap areas and blank areas restricts the performance of the depth-only methods. To address this issue, we proposed the Coupled U-Net (CU-Net) model that is enhanced by a confidence-based outlier removal algorithm. Our method achieves the state-of-the-art performance on the KITTI leaderboard and has a good generalizability under different depth densities and various conditions. Meanwhile, our work advocates that a feasible way to improve the results in the overlap and blank areas, such as designing more complex algorithms or introducing data from other modalities, may be the focus of future research.

\bibliographystyle{IEEEtran}
\bibliography{IEEEabrv,references_abrv}

\section*{supplementary materials}

This part is the supplementary materials for ``CU-Net: LiDAR Depth-only Completion with Coupled U-Net''.
In \secref{sec:sec1}, we show qualitative experiments on the DDAD dataset.
In \secref{sec:sec2}, we describe some typical examples of a continuous video sequence.

\renewcommand\thesection{\Alph {section}}

\setcounter{equation}{0}
\setcounter{section}{0}
\setcounter{subsection}{0}
\renewcommand{\theequation}{A.\arabic{equation}}
\renewcommand{\thesubsection}{A.\arabic{subsection}}

\section{Qualitative experiments on the DDAD dataset}
\label{sec:sec1}

The Dense Depth for Autonomous Driving (DDAD) dataset~\cite{DDAD} is a new autonomous driving benchmark from Toyota Research Institute for long range (up to 250m).
The dataset contains monocular color images and accurate ground-truth depth generated from a high-density, long-range Luminar-H2 LiDAR sensor.
Unlike the KITTI dataset captured under good lighting and weather conditions, the DDAD dataset contains scenes with different lighting conditions (such as morning, sunset, and overcast) and weather conditions (for example, rain).  
We conduct experiments under various conditions in DDAD to verify the generalizability of our model.
In experiments, we crop the input depth maps to $512\times1216$.

\begin{figure}[!h]
	\centering
	\includegraphics[width=0.5\textwidth]{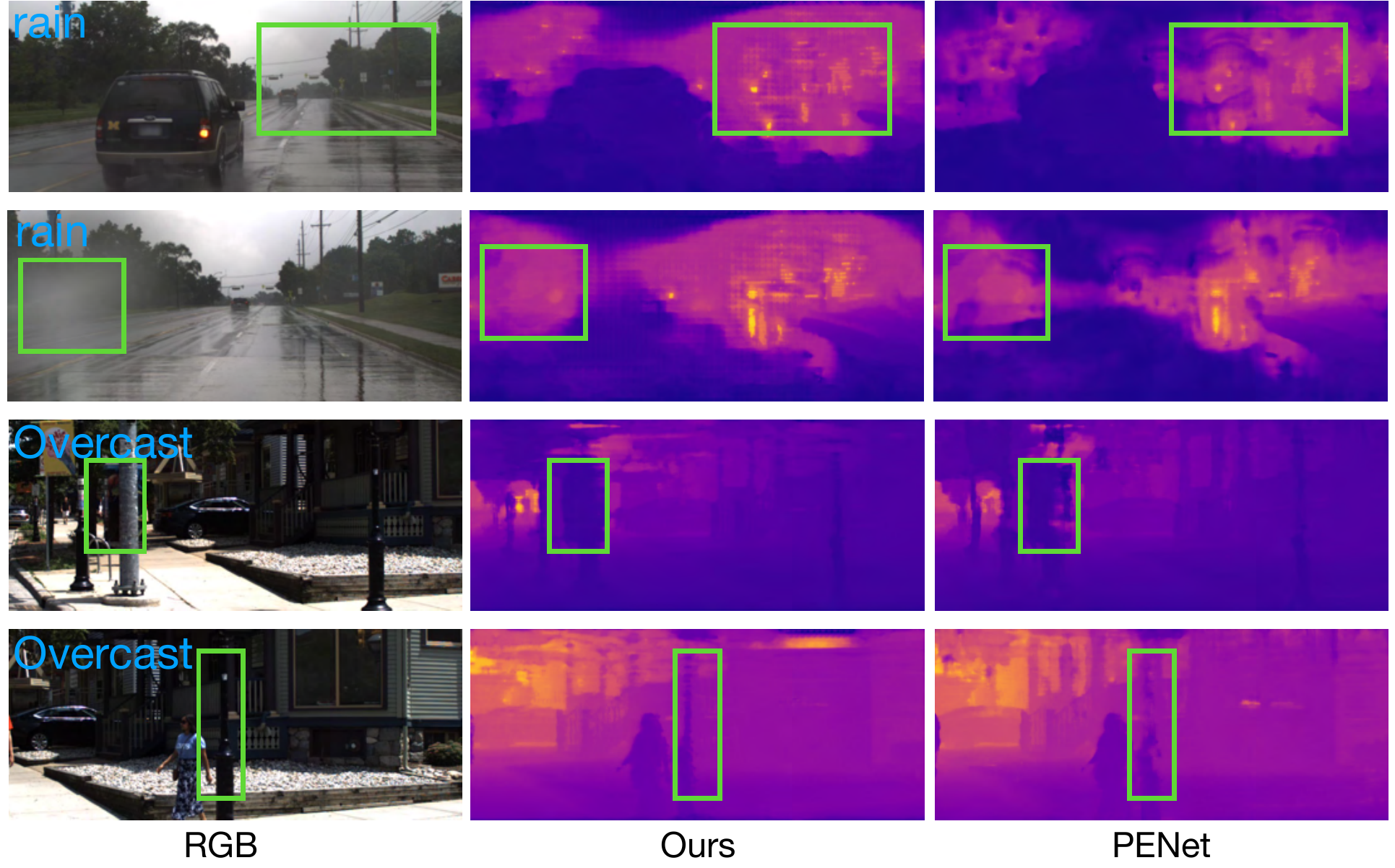}
	\caption{Typical examples of our model and the PENet on the DDAD dataset.}
	\label{fig:ddad_result}
\end{figure}

To verify the generalization of our model in various conditions and datasets, we use the model trained on the KITTI dataset to directly test on five sequences with different lighting and weather conditions of the DDAD dataset.
The experimental setup is difficult because there are many differences between training and inference data.
We compare our method with the state-of-the-art RGB-guided method PENet~\cite{HuWLNFG21}.
The PENet model is provided by the open-source project, whose RMSE is 757 in the KITTI validation dataset.
Fig.~\ref{fig:ddad_result} shows some visualization results of our model and the PENet on the DDAD dataset.
The first two rows of samples are collected on rainy days, and the last two rows of samples are collected on overcast days.
We observe that for adverse weather like rain, the dense depth map predicted by our method is more complete and clear compared to PENet.
Meanwhile, for the overcast samples, the object boundaries in our dense depth maps, such as the edges of columns and street lights, are also sharper.

\section{Experimental results of a continuous video sequence}
\label{sec:sec2}

To show the experimental results more clearly, we provide some typical examples of a continuous video sequence.
The results include the sparse depth map, the sparse depth map after removing outliers, the confidence map predicted by the local U-Net and the global U-Net, our results, and the ground truth.
The results show that the confidence images predicted by the local U-Net have high confidence values in the normal areas. In contrast, they have low confidence values in the overlap areas and blank areas. The confidence images predicted by the global U-Net are the opposite.
The video is released at \url{https://youtu.be/WAQouCouw5E}.
	
\end{document}